\documentclass[10pt, a4paper]{article}

\usepackage[final]{lrec2026} 
\usepackage{algorithm}
\usepackage{algpseudocode}

\title{ReX-GG: a LLM Ensemble Pipeline for Relation-extraction and Graph Generation  }

\name{Giacomo Magnifico, Eduard Barbu} 

\address{
Institute of Computer Science, University of Tartu \\
Tartu, Estonia \\
\texttt{\{giacomo.magnifico, eduard.barbu\}@ut.ee}
}

\abstract{
Current LLM ensemble frameworks focus on multi-step setups with additional modules for answer ranking, often opting for token and span analysis rather than structured outputs, leading to heavyweight architectures with potential fail states along the pipeline. Faster, lighter solutions are more vulnerable to hallucination propagation and can lack output control in more complex pipelines. This paper proposes a customisable, lightweight ensemble workflow of coordinated Large Language Models that leverages JSON-structured outputs and anonymous peer-review ranking to mitigate hallucinatory outputs and single-model failure points. The pipeline is demonstrated on a relation extraction task applied to popular science articles in English, targeting four ontologically-grounded relation types (strong causation, weak causation, contrastive, and compositional), with semantic node canonicalisation and interactive, colour-coded HTML causal graphs as the final output. Performance is evaluated through an anonymous user study, achieving an average perceived accuracy of 0.778 against a human-annotated gold standard. The modular architecture supports flexible deployment across both API-based and in-house LLM setups, and the full framework is released under an open license to foster reproducibility and collaborative research.
 \\ \newline \Keywords{graph, human survey, lightweight framework, LLM ensemble, relation extraction, structured output} }

\begin{document}

\maketitleabstract

\section{Introduction}

Since the advent of OpenAI and the permeating use of Large Language Models (LLMs) in everyday life and academia, the ease of access to state-of-the-art models and diffusion has led to new avenues of research; from testing the limits of these models to mitigating their shortcomings, exposing their faults and tuning their performance. As expected, the use of individual models has led to the development of LLM ensembles, tailored to achieve more robust performance and to leverage perceived differences among these models \citep{jiang-etal-2023-llm}. Other ways to enhance the reliability and robustness of their output have been found at the intersection of formal structure and natural language processing (NLP), in the form of information-based graphs, with examples of knowledge-graph-informed architectures that leverage the power of ontologies and formal structure \cite{zhou-etal-2025-reflection, zhu-etal-2025-knowledge}. Within this specific niche, it is possible to find widely different research avenues - from graph-based LLM ensemble performance evaluations \cite{he-etal-2025-llm} to analyses of LLM inner workings via graph representation of their chain of thought \cite{zhiqiang-etal-2025-judge}. The production of knowledge graph outputs seems to be less explored, with  \citet{le-etal-2025-docie} and \citet{li-etal-2025-beyond-llms} as examples, although only the former explores ensembles and structured outputs, while the latter focuses on LLM tuning.

To foster further research in this niche and to provide an accessible tool to the research community, this paper presents a \textbf{lightweight example of an information-extraction ensemble workflow, tailored for relation extraction and interactive graphs}. The proposed workflow can be deployed with a customisable selection of coordinated models, delivering robust performance and a reduced risk of hallucinatory or malformed output thanks to the peer-reviewed ranking of schema-enforced JSONs. The practical application of the pipeline targets English popular science articles, with performance tested via an anonymous survey to provide public-informed evaluation of the relation-extraction framework; anecdotal evidence shows a preference for colour-coded graph edges representative of the evaluated bonds. The secondary advantages of this work can be identified as the following:
\begin{itemize}
    \item simple overarching structure with flexible end use;
    \item built-in mitigation of hallucination propagation;
    \item reduced risk of malformed outputs failure-chains;
    \item fast deployment as a modular piece in larger architectures.
\end{itemize}

Details regarding the structure and behaviour of the ensemble are provided in Section \ref{methods-models}, along with the ontology and details regarding the nature of the relations chosen for extraction; further information related to the canonicalization of the extracted pairwise relation triplets and development of the graph structure are given in Sections \ref{methods-embed} and \ref{methods-graph}. In Sections \ref{discussion} and \ref{conclusions}, we discuss the results of our user study presented in Section \ref{results}, expand on the key findings of this paper, and outline potential future work. We reserve Sections \ref{limits} and \ref{ethics} to discuss the shortcomings and limitations of our scope and to address potential concerns about the ethical standards of this work.

\section{Related Works}

Multiple works deploy different sets of LLMs as ensembles with variable cohesion for a multitude of purposes: from \citet{xu-etal-2025-hit}, which evaluates a 7B selection of models and enhances decision-making through token span analysis, to the more complex database-oriented tasks focusing on tabular data QA seen in \citet{bujnowski-etal-2025-samsung}, to striving for the optimization of the ensemble with additional framework refinement as proposed by \citet{tekin-etal-2024-llm}.

Related to our work are the concepts and implementation of the "LLM forest" found in \citet{he-etal-2025-llm}, which consists of a graph-\textit{informed} ensemble of models to achieve higher performance in data inputation for subsequent downstream tasks; the performance of graph-informed processes is further investigated in \citet{zhiqiang-etal-2025-judge}, alongside the Explainable Graph Language Model framework. Graph-\textit{represented} decisions, which are closer to the scope of this paper, are explored in \citet{xiong-etal-2025-mapping} as a means to clarify the reasoning process of LLMs, converting large freeform text chain-of-thoughts (CoTs) into a readable and measurable graph to explain the causal bonds; the difference in scope lies within model-reasoning applications compared to extractive tools for information transmission. While the output of LLMs is converted into graph form, and the enforced structure within responses is a shared property, the work presented in \citet{wu-etal-2025-medical} makes use of external repositories and ultimately outputs natural language text based on knowledge graphs.

The two papers that share most of the background with the ensemble of our proposed work are \citet{parfenova-pfeffer-2025-samebutcode} and \citet{le-etal-2025-docie}; while the former proposes a two-staged ensemble with the aid of a moderator to be deployed for inductive coding, with an emphasis on the use of LoRA finetuned architectures, the latter provides a two-stage pipeline with a smaller zero-shot ensemble. Despite the structural similarities, our work differs both in scope and use case: the focus on causal relations and the use of few-shot prompts distances it from \citet{le-etal-2025-docie}, and the preferred deployment as an intermediary between freeform text and HTML graphs falls further away from \citet{parfenova-pfeffer-2025-samebutcode}'s scope.

Lastly, what could be considered a precursor for this work is presented in \citet{li-etal-2025-beyond-llms} as a single-model pipeline that specifically produces causal graphs from narrative text; although both papers share similar goals, the core difference lies in the use of finetuned BERT architectures and summarisers to overcome the performance of base LLMs, turning the pipeline into a robust but heavy-weight solution.

The reviewed literature reveals that while LLM ensembles have grown increasingly capable, their complexity has scaled accordingly, often requiring additional ranking modules, finetuned architectures, or external knowledge repositories. Works that produce graph-structured outputs tend to focus on model reasoning transparency rather than information transmission, and those that do target relation extraction either rely on heavyweight finetuned pipelines or limit themselves to single-model setups. No existing work combines a lightweight heterogeneous ensemble with schema-enforced structured outputs, ontologically-grounded relation types, and interactive graph visualisation in a single deployable framework. This paper directly addresses that gap, proposing a \textit{modular, robust yet lightweight solution for everyday information extraction that prioritises accessibility and ease of deployment without sacrificing output reliability}.

\section{Pipeline Structure}

\subsection{Model Ensemble}
\label{methods-models}

\begin{figure*}[ht]
    \centering
    \includegraphics[width=0.97\linewidth]{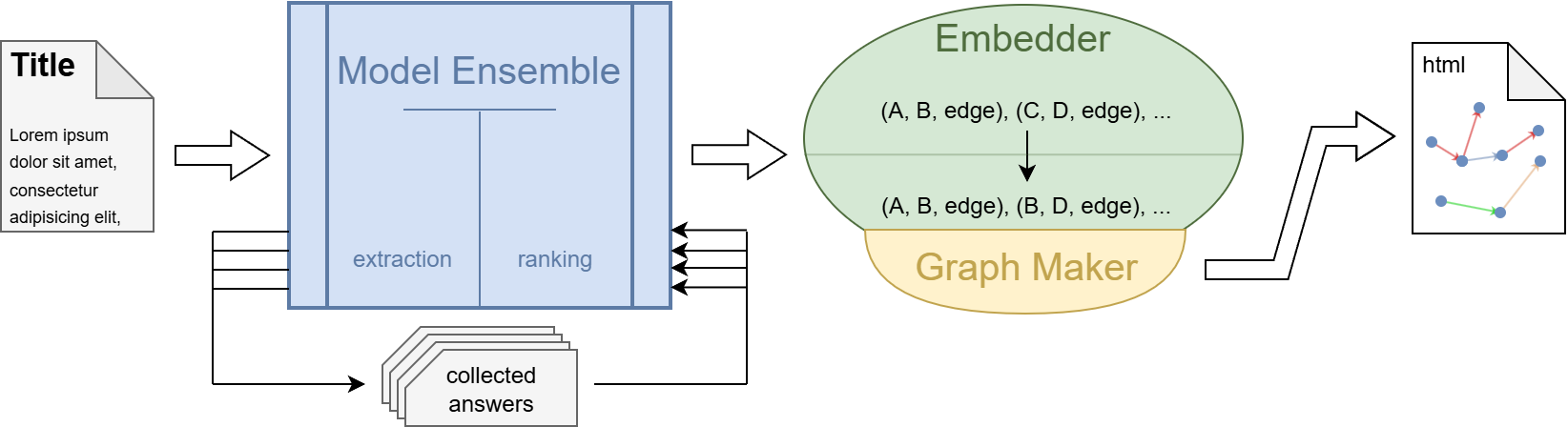}
    \caption{Schematic representation of the pipeline presented in the paper: a text file is fed to the LLM ensemble, where relation extraction is performed, and the best output is decided through anonymous peer-ranking; the extracted relation triplets iterate through an embedding model to improve graph structure and assimilate nodes with semantic similarity. The final result is a causal graph in HTML format.}
    \label{fig:pipeline-long}
\end{figure*}

\textbf{Relation Extraction}. After receiving input from the user with the name and extension of the file to be processed, the read information is stored and sent to each model in the ensemble for the preliminary relation extraction. Our chosen focus is on narratively-driven direct pairwise relations, examples of which can be found both in scientific-literature-based tasks \cite{le-etal-2025-docie} and narrative text \cite{li-etal-2025-beyond-llms}; the relation types we focus on are based on the works by \citet{Ross_2025} and \citet{magnifico-2025-automated}, among others, and are defined below.
\begin{itemize}
    \item \textbf{Strong Causation}. Explains a phenomenon by specifying explicit causal mechanisms or multiple steps in a causal chain. The underlying process is made clear. \textit{“Your lack of sleep is making you clumsy. That’s why you spilled the milk.”}
    \item \textbf{Weak Causation}. Explains through correlational or probabilistic relationships between variables without specifying the underlying mechanism. Often involves associations, tendencies, or indirect influences. \textit{“Diseases with uncertain causes were about 50\% more likely to attract religious or magical treatments.”}
    \item \textbf{Contrastive}. Explains a phenomenon by highlighting differences between alternative conditions or cases (A vs B). \textit{“In the treatment group, 70\% recovered; in the placebo group, only 30\% recovered.”}
    \item \textbf{Compositional}. States what something is made of, contains, consists of, or is defined as (whole/entity → part/definition). \textit{“Protons and neutrons are combinations of even tinier particles, called quarks.”}
\end{itemize}
To maximise the potential of the extraction pipeline, further rules for the two extremes of the pairwise relations are introduced. First, each leftmost extreme (the future \textit{head} node) is defined as a \textit{[concept, cause, outcome A, entity]}, while the rightmost extreme (the future \textit{tail} node) is defined as a \textit{[concept, effect, outcome B, constituent]}. Second, each extracted extreme must be a \textit{minimal span of words}, preferably a \textit{nominal phrase} - as it is the minimal identifier for a concept, be it a subject or an object in a sentence - and with each pronoun/anaphora resolved to its \textit{explicit referent}. Third, rewriting the original text is preferable to quoting long sentences directly. A possible example of the extracted relations for the paragraph 
\begin{quote}
    \textit{``Of course your hands are all jittery, you had a big gulp of coffee earlier! Maybe that's why you're dropping stuff?''}
\end{quote}
would be \textit{[``big gulp of coffee'', ``jittery hands'', strong causation]}, and \textit{[``jittery hands'', ``dropped stuff'', weak causation]}.
After lengthy tuning, the final set of instructions follows the best practices for prompt engineering and can be summarised with this outline:  
\begin{itemize}
    \item task definition;
    \item output schema example;
    \item rules for schema completion;
    \begin{itemize}
        \item rules for node extraction;
        \item rules for edge extraction;
    \end{itemize}
    \item set of input-output examples
\end{itemize}

As a reliable method of enforcing the schema structure given with the prompt, the choice of \texttt{structured\_output} settings is enabled; the resulting output is in JSON format as a single-entry Python dictionary, with the word "document" as the \textit{key} and a list of smaller dictionaries as the \textit{value}. The list formatting allows for the iterative use of dictionaries with the same shared keys ("bond", "span 1", "span 2") associated with different values, corresponding to the extracted pairwise relations and relation extremes. In this section of the pipeline, there is no interaction among the models, and each output is produced independently. All the JSON-structured outputs from this first pass through the ensemble, representative of the relation bonds extracted from the input document, are then passed along in the pipeline.\\

\textbf{Model Peer Ranking}. As the outputs of the first ensemble pass have been stored separately for each individual model, they are anonymised to avoid both potential self-preference bias from any architecture and tuning-induced preferences present in the training data. To facilitate easy post-processing de-anonymisation, each model is assigned a corresponding letter from the English alphabet, starting with A. The anonymised outputs are then given to each model in the ensemble, along with the original document and a reminder of the classification rules, with the explicit task of assigning a 10-point grade to each.

Once more, these outputs are forced into a JSON structure and must provide the following dictionary entries:

\begin{itemize}
    \item \textit{answer :} string, label of the ranked answer (e.g. A, B, C);
    \item \textit{points :} integer, points assigned to the ranked answer (0 to 10);
    \item \textit{thinking :} string, providing a reasoning behind the score of the ranked answer.
\end{itemize}.

The points for each answer label are stored and summed as each model in the ensemble goes through this process, forming a "leaderboard" with the comprehensive rating of each anonymous output. Afterwards, the answers are de-anonymised, and the user is informed of \textit{how high each model's output scored, along with the one chosen as the most accurate}. The best-voted output is then sent along the pipeline. \\

Of relevant mention is the use, through the entire ensemble pipeline, of continuously updated \texttt{query\_logs} and \texttt{error\_logs}; not only is the entirety of the content generated by the ensemble safely stored in a separate text file, but each potential break in the sequence due to unstructured outputs is recorded in a dedicated log file. From the input given to the ensemble to the final structured output, everything is stored for maximum clarity of use.

The models used for the setup presented in this paper have been deployed through API calls via the OpenRouter platform\footnote{\url{https://openrouter.ai/}}, with an approximate cost of £10 for the entirety of the multi-stage process. One model for each major corporation has been chosen based on popularity and performance ratings on aggregator websites as of February 2026. The final ensemble of LLMs is comprised of GPT 5.2 \citeyearpar{openai}, Claude Sonnet 4.5 \citeyearpar{claude}, Gemini 3 Flash \citeyearpar{google-gemini}, Deepseek 3.2 \citeyearpar{deepseek}, Llama 3.3 (70B-instruct) \citeyearpar{meta-llama}, Nova 2 Lite V1 \citeyearpar{amazon-nova}, Grok 4.1 Fast \citeyearpar{grok}; all of the outputs are set to \texttt{structured}, with a 4000 token limit and the same fixed seed for the entire process to keep consistency as high as possible. 

\subsection{Embedding Model}
\label{methods-embed}

After the LLM ensemble, the second block of our pipeline prepares the best-voted answers (as lists of dictionary-encoded lists of semantic triples) for later graph representation. As the semantic triples are stored with a high likelihood of non-identical text strings for similar concepts (e.g. ``a coffee'', ``coffee'', ``nice cup of coffee''), a graph derived by the raw list would be scattered and represent only two-node relations; with no changes, the compositional properties of causal bonds would be lost, as well as any other defined relation type that spans multiple concepts. To achieve a cohesive structure and minimise fragmentation, we turn to semantic embedding as a lightweight yet effective form of control.  Each list of dictionary-encoded relation pairs is processed as detailed in Algorithm \ref{alg:cap} before moving forward in the pipeline.

\begin{algorithm}
\caption{Semantic Triples Node Assimilation}\label{alg:cap}
\begin{algorithmic}

\State $inputs \gets$ list of dictionaries
\State $relations = [$   $]$
\For{$i$ in $inputs$}
\If{$i[relation] \neq$ none}
  \State $pair \gets (i[A], i[B], i[relation])$
  \State $relations \gets relations + pair$
\EndIf
\EndFor
\For{$i=1$ to $i = \#relations$} 
\State $x \gets relations[i-1][1]$
\State $y \gets relations[i][0]$
\State $X, Y = embedder.encode(x,y)$
\If{$embedder.similar(X,Y) \geq value$}
\State $y \gets x$ 
\EndIf
\EndFor

\end{algorithmic}
\end{algorithm}

The first half of the process shows how the inputs received from the LLM ensemble are converted from dictionary encoding into \textit{tuple} object types, all while keeping the RDF structure \[ A \rightarrow relation \rightarrow B \] with possible examples being \[ coffee \rightarrow strong \ causation \rightarrow \ jittery \ hands \] \[ ´jittery \ hands´ \rightarrow weak \ causation \rightarrow ´drop \ stuff´. \]
An immediate change in the conversion is the substitution of the relation types mentioned in Section \ref{methods-models} with the colour associated with them. Since the final output of the pipeline has to be intuitively understandable and easy to read, we opt for the colour-coding \textit{[red, orange, cyan, green]} for \textit{[strong causation, weak causation, contrastive, compositional]}. Thus, the two examples above would be converted into \[ coffee \rightarrow red \rightarrow jittery \ hands \] \[ jittery \ hands \rightarrow orange \rightarrow drop \ stuff. \]

Worth noting is that each pairwise relation is stored in a \textit{A-B-relation} order, as a popular format for graph information storage is the \textit{(head, tail, directed edge)} notation.  

Presented in the second half of Algorithm \ref{alg:cap} is our solution for graph sparcity: an iteration over the list of stored semantic triples using an embedding model to achieve a rudimentary but efficient form of multi-node relation canonicalization. As subsequent triples are iterated through, the \textit{tail} of the previous one is compared with the \textit{head} of the following through vector embedding and similarity measure between the embeds; if the semantic similarity is higher than a chosen threshold, the two concepts are assumed to be syntactic variations of the same semantic term. If that is the case, the two nodes become one and the same, transforming a pair of arcs \[A \rightarrow relation \rightarrow B\] \[C \rightarrow relation \rightarrow D\] into \[A \rightarrow relation \rightarrow B\] \[B \rightarrow relation \rightarrow D\] which represent the assimilated triplet composition \[A \rightarrow relation \rightarrow (B \rightarrow relation \rightarrow D).\] We deploy \texttt{Jasper-Token-Compression-600M} \citep{zhang2025jasper} as a semantic embedding model, given its superior performance relative to near-state-of-the-art language models.

\subsection{Interactive Graph Maker}
\label{methods-graph}

\begin{figure}[h!]
    \centering
    \includegraphics[width=\columnwidth]{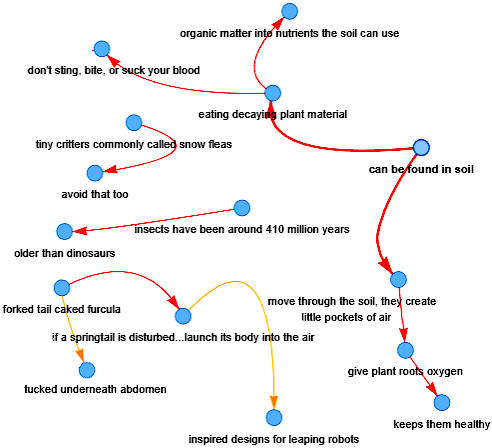}
    \caption{A full graph output from our pipeline, derived from the article "Snow fleas use their tails to jump around the ice", available at \href{https://www.popsci.com/environment/snow-fleas-jumping/}{this link}.}
    \label{fig:graph-sample}
\end{figure}

Given the input from the embedding algorithm, we populate a \texttt{NerworkX}\footnote{\url{https://networkx.org/en/}} directed graph with given dimensions of 1080p $\times$ 1080p. The directed graph is then converted into a \texttt{PyVis}\footnote{\url{https://pyvis.readthedocs.io/en/latest/index.html}} object to allow for better interactivity, and it is exported as an HTML file - akin to the sample provided in Figure \ref{fig:graph-sample} as an image. We chose this specific format because it enables a more hands-on approach for the end user, allowing them to move nodes around, enable/disable graph physics, highlight nodes and vertices, and so on.

\section{Human Evaluation}
\label{results}

To avoid authorship bias in the effectiveness and correctness of the graphs generated by the pipeline, an anonymous survey was distributed to paid volunteers recruited via the \textit{Prolific} web app\footnote{\url{https://www.prolific.com/}}. Two graphs were annotated by the authors to serve as a human-annotated gold standard baseline, and then evaluated by two dedicated groups of 3 anonymous testers. The remaining 18 articles in the set were processed by the pipeline and each was evaluated by 6 testers. The scoring method was as follows: a four-point evaluation of the usefulness of the graph; a five-point evaluation of the accuracy of the information provided in the graph; and a final four-point score for the tester's confidence in their scores. Although the perceived usefulness of the graphs was not considered in the current evaluation, the testers' confidence was used to compute a weighted average across annotator groups for each article. The scores for the accuracy of author-annotated graphs served as a baseline for comparing the accuracy of the remaining data, as they represented the "ideal accuracy" for manual annotators.

\begin{table}[h]
\centering
\begin{tabular}{llll} \hline
   & \textbf{Accuracy} & \textbf{Length} & \textbf{$\times\sigma$}  \\ \hline
A1  & 0.807 & 2342  & 0.2 \\
A13 & 0.939 & 2374  & 1.1 \\
A14 & 0.903 & 2444  & 0.9 \\
A7  & 1.000 & 2603  & 1.6 \\
A10 & 0.745 & 3062  & 0.2 \\
A12 & 0.880 & 3245  & 0.7 \\
A5  & 0.550 & 3270  & 1.6 \\
A18 & 0.521 & 3709  & 1.8 \\
A3  & 0.880 & 3895  & 0.7 \\
A15 & 0.953 & 4253  & 1.2 \\
A6  & 0.623 & 4426  & 1.1 \\
A4  & 0.790 & 4764  & 0.1 \\
A8  & 0.835 & 4967  & 0.4 \\
A16 & 0.831 & 5215  & 0.4 \\
A11 & 0.636 & 5950  & 1.0 \\
A17 & 0.815 & 9078  & 0.3 \\
A2  & 0.632 & 9542  & 1.0 \\
A9  & 0.670 & 10321 & 0.8 \\
    &       &       &     \\
\textit{Avg.} & \textit{0.778} & \textit{4748} & \textit{1}   \\ \hline
\end{tabular}
\label{tab:graph-accuracy}
\caption{Scores for the perceived accuracy of each graph, normalised against the human-annotated gold standard, with the distance from the mean measured in standard deviations.}
\end{table}

\begin{figure*} [ht]
    \centering
    \includegraphics[width=0.9\linewidth]{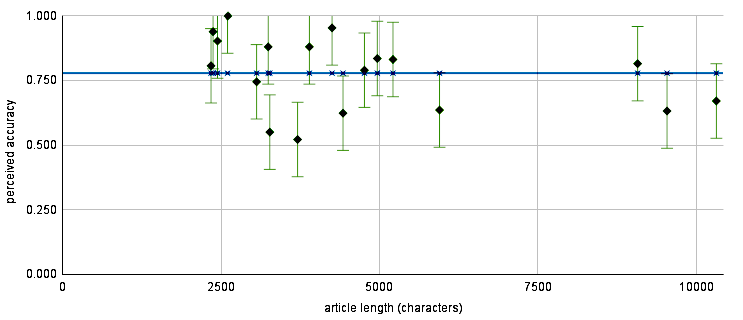}
    \caption{Performance of the graph produced by our pipeline according to the anonymous survey. The averaged accuracy scores for each article are represented as filled $\diamond$, along with the line marking the mean value. }
    \label{fig:graph-accuracy}
\end{figure*}

Divided by original document and in increasing length order, the scores for each graph's perceived accuracy and standard deviation steps from the average are shown in Table \ref{tab:graph-accuracy}. A visual representation is available in Figure \ref{fig:graph-accuracy}, along with the projected mean of the perceived accuracy and a \(\pm \sigma\) representation. The furthest outliers in the data (A5, A7, A18) are still within 2 $\sigma$ from the mean, and the majority of values are within one $\sigma$.

\begin{figure*} [ht]
    \centering
    \includegraphics[width=0.9\linewidth]{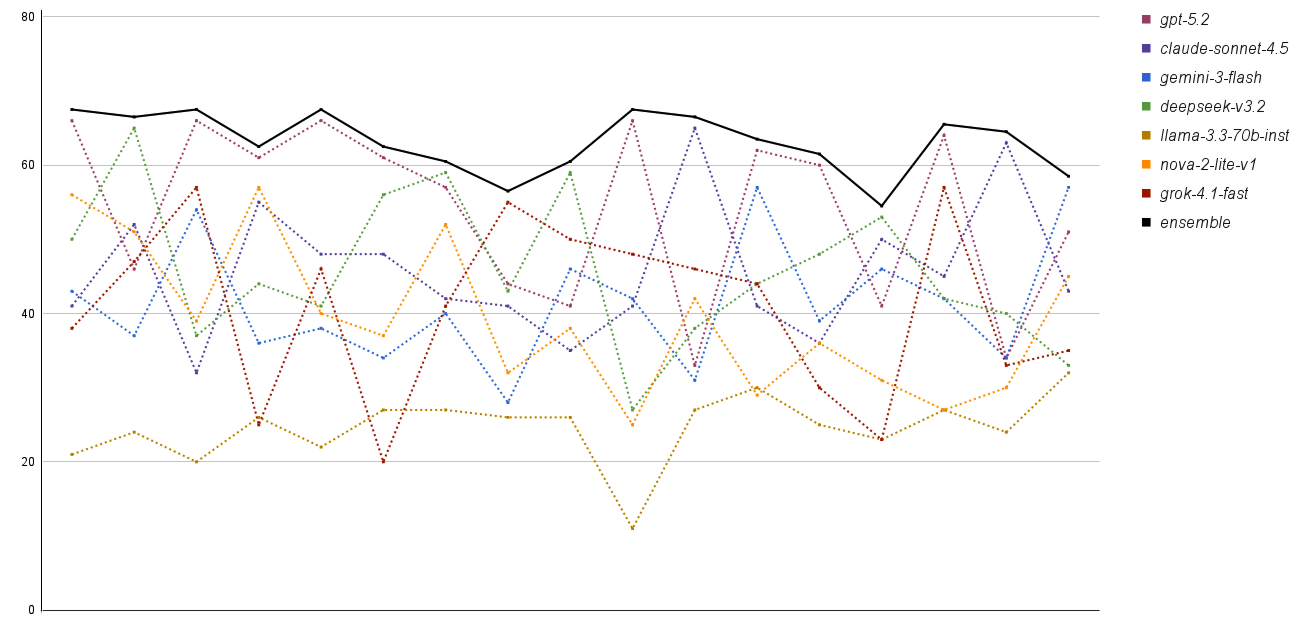}
    \caption{Ranking of our ensemble scores compared to the performance of the individual model, ranging from a minimum of 0 to a maximum of 70 points, over the articles evaluated.}
    \label{fig:graph-semble}
\end{figure*}

To provide a visual representation of the robustness of our proposed solution, Figure \ref{fig:graph-semble} presents the average grading for each individual model (dotted lines) across the selection of articles, with the model ensemble pipeline as a comparison (full line).

\section{Performance Analysis}
\label{discussion}

The processed results from the anonymous survey can be helpful in defining the key properties of the deployed pipeline, ranging from the relation between performance and document length and the perceived accuracy of the graphs to the robustness of the ensemble.

\textbf{Ensemble vs Single-model Rank}. As shown in Figure \ref{fig:graph-semble}, the variability in the performance of each single LLM presents a wide margin of potential error that can lead to inaccurate graphs and poorly-formed outputs. Even the best-performing model is shown to have instances of rankings lower than 30 points, and the less-performing models at times rise to higher scores. Overall, the ensemble ranking leads to more stable, robust performance.

\textbf{Document Length vs Accuracy}. Based on the results shown in Table \ref{tab:graph-accuracy} and the distribution presented in Figure \ref{fig:graph-accuracy}, we can conclude that there is no direct relation between document length (measured in characters) and perceived accuracy of the relative graph. Both the Table and the Figure show that the highest accuracy and the lowest are both within the initial range (2000 to 5000 characters); it could be speculated that the accuracy scores show a diminishing trend as the documents become longer, but a more balanced evaluation would be that the variance between results lowers with longer documents. 

\textbf{Human-rated Accuracy}. Taking into consideration the mixed nature of the Accuracy score, indicative both of completeness compared to the original text (e.g. number of core relations reported in graph) and accuracy of the information presented (e.g. two elements put in correlation rather than contrast), our pipeline produces high-quality outputs with near-human annotation peaks and an average of \textit{0.778}.
All of the human ratings for the interactive graph outputs are within 2 standard deviations from the mean, with the majority (66\%) being within 1 standard deviation. Even without the removal of the largest outliers in both directions, the limited span of distribution for the ratings supports the assumption of an effective pipeline. 

\textbf{Robustness}. Due to the nature of the ensemble architecture, each of the models is forced to produce a JSON output, and each part of the content-generation and recording sequence carefully looks for formatting errors; as each output is stored independently and given anonymously for peer-review ranking to each model, the pipeline avoids the pitfalls of hallucination propagation and unusable outputs. The anonymous nature of the ensemble rankings allows for personalised use with a \textit{single LLM ensemble setups} for even lighter-weight scenarios, and it is adaptable to both in-house and API LLM usage. 

\section{Conclusions}
\label{conclusions}

In this paper, we have presented a lightweight pipeline for information extraction that leverages an ensemble workflow of seven independent models, coordinated to reliably mitigate hallucinatory outputs and malformed graph generation. Our framework achieves robust performance by leveraging JSON-structured outputs and independent model-peer-review ranking strategies. When the deployed pipeline is tested on a selection of English popular science articles, the reported performance is close to the human-annotated gold standard, with an average accuracy of the provided information of 0.778 - measured by aggregated scoring from anonymous annotators. The outputs are presented to end users as interactive, colour-coded HTML files that facilitate immediate, intuitive understanding.

Anecdotal evidence supports the ease of use and deployment of the proposed pipeline even by less-specialised users, and the modular nature of the ensemble allows for highly customisable settings; possible implementations that benefit from the anonymous peer-review model-ranking include free and open source solutions, such as in-house single-LLM ensembles with multiple-persona structures, or heterogeneous ensembles that can benefit from an array of different-scale models.

Future work is necessary to reach higher performance and more interpretable results, as reports from anonymous users in the study suggest that a central topic node and a specific top-down or left-right directionality would be beneficial for graph understanding. Further testing is also necessary to verify the robustness of our pipeline across different settings and contexts, aiming for higher generalisation and a wider selection of relation types; a potential addendum could be investigating input multimodality to determine the best format for maximal accuracy. Finally, developing a simpler software solution that allows non-command-line execution of the pipeline would improve accessibility and ease of use for the general public.

All the software presented in this paper is freely available at our GitHub repository\footnote{\url{https://github.com/gima9552/ReX-GG}} to foster cooperative research and an open-science mindset in the academic community.

\section{Limitations}
\label{limits}

Since our paper focuses on presenting a lightweight alternative to heavier information-extraction pipelines, some limitations regarding performance and size are bound to the nature of the work. The current workflow has been tested with shorter documents, which are easier to read and evaluate for the anonymous testers; consequently, proper testing with sizeable documents (e.g. entire scientific research papers) falls outside of the current scope despite it being, perhaps, the more interesting application for our pipeline. For the sake of simplicity and customisation, aside from the processed data and the log information, all the information flow is strictly online, as the models are called through API, and a true analysis of the reasoning process for LLMs is not available. Anecdotal evidence also suggests that the output graphs would benefit from a central node with the general topic of the input document as the origin, with radial orientation to improve comprehension. Lastly, the human evaluation is more qualitative than quantitative, as the number of evaluated documents is on the lower end of the spectrum. 

\section{Ethical Considerations}
\label{ethics}

The reliance on LLMs to extract information from texts and present it to the public can inherit data-induced bias and carry the harm of potential misinformation. Our framework mitigates the propagation of hallucinatory outputs by leveraging the ensemble-based ranking, but it is not guaranteed to have a completely faithful output in every single case; therefore, the recommendation is to always introduce human supervision in the loop, and generally defer judgment to the user. The evaluation done with anonymous testers was conducted with no risk of harm and within safe work regimes, and the payment was above the platform's minimum wage standards. To the best of our knowledge, the work presented in this paper does not pose any significant risk of harm in itself.

\section{Acknowledgements}
This work was supported by the project “Terminology-Aware Machine Translation for Accessible Science” (TaMTAS, MOB3ERA10), funded by the Estonian Research Council under the Mobilitas 3.0 ERA-NET programme and co-funded by the European Union under the CHIST-ERA Call 2025 “Science in Your Own Language.”

\section{Bibliographical References}

\bibliographystyle{lrec2026-natbib}
\bibliography{lrec2026-example}

\end{document}